# Exploring the impact of weather on Metro demand forecasting using machine learning method

Yiming Hu,[1,*] Yangchuan Huang,[2] Shuying Liu,[3] Yuanyang Qi,[4] and Danhui Bai[5]

[1] Qiushi Honors College, Tianjin University, Tianjin, 300354, China
[2] School of Traffic and Transportation, Beijing Jiaotong University, Bejing, 100091, China
[3] Maynooth International Engineering College, Fuzhou University, Fuzhou, Fujian Province, 350100, China
[4] Department of Automation, Harbin Institute of Technology, Harbin, Heilongjiang Province, 150001, China
[5] Department of Foreign Languages, Beijing Forestry University, Beijing, 10083, China

* Email: hymglitter@tju.edu.cn

## Abstract

Urban rail transit provides significant comprehensive benefits such as large traffic volume and high speed, serving as one of the most important components of urban traffic construction management and congestion solution. Using real passenger flow data of an Asian subway system from April to June of 2018, this work analyzes the space-time distribution of the passenger flow using short-term traffic flow prediction. Stations are divided into four types for passenger flow forecasting, and meteorological records are collected for the same period. Then, machine learning methods with different inputs are applied and multivariate regression is performed to evaluate the improvement effect of each weather element on passenger flow forecasting of representative metro stations on hourly basis. Our results show that by inputting weather variables the precision of prediction on weekends enhanced while the performance on weekdays only improved marginally, while the contribution of different elements of weather differ. Also, different categories of stations are affected differently by weather. This study provides a possible method to further improve other prediction models, and attests to the promise of data-driven analytics for optimization of short-term scheduling in transit management.

## 1. Introduction

Urbanization in China is accelerating with the population growth in metropolis like Hong Kong, straining the city transport system. Eco-friendly transformation is required to decrease the carbon mission per capita. In this context, metro system with transport sharing, higher security, and lower carbon mission is desirable. Accurate and efficient prediction of passenger flow is required for emergency response as well as travel time and transport pattern decision for the public. Such prediction would further enable smart train scheduling and help public transport resources meet travel needs.

Many studies have been investigating on methods to forecast passenger flow, methods of which can be divided into three types, statistical models, classical machine learning models and neural network models[1]. Among all the statistical models, autoregressive integrated moving average (ARIMA) is the most popular one. It establishes mathematical model based on the previous values. In 1970s, Ahmed et al.[2] firstly applied ARIMA on short-term traffic flow prediction problem. Since then, the model has been continuously improved by adding seasonal factors[3] and so on. However, ARIMA only uses information of or conducted from previous values, while it is not suitable to reveal the nonlinearity of passenger flow.

Due to the above drawbacks of classical statistical models, lots of studies have focused on machine learning models. Machine learning models learn from the data and determines parameters automatically[4]. Some classical machine learning models, such as random forest, bagging regression, and support vector regression (SVR), were proposed and applied to passenger flow forecasting. Liu et al. applied random forest to compare availability of models on different types of date[5]; Su et al. used SVR and adaptively adjusted the model according to new data collected (namely incremental SVR) to predict short-term traffic flow[6]. Neural network, which can extract feature automatically[7,8], is one of the most popular machine learning models nowadays. Dougherty et al. first applied it to traffic flow prediction[9]. Guo et al. simply apply MLP to predict the outgoing traffic of the multiple directions for a particular crossroad[10]. This model is also used by Belay Habtie et al. (2015). More prevalent deep neural network-based models including CNN[11], LSTM[12], GRU[13] or their combination[14] are applied to predict passenger flow for further improvements of time-series related variables. However, despite the enhancement brought by mathematical models and number of iterations, most of the work only used passenger flow data by time to predict future flow. Some of them categorize dates to account for different traffic paradigms on holidays and weekdays, but few works paid attention to environmental data, which may also significantly affect the travel decision of passengers[15], and thus affect the passenger flow. Also, while lots of works introduced additional variables like types of date dummy or environmental variables to their models, few paid attentions that to what extent the additional variables help to improve the model.

As for the impact of weather on passenger flow, lots of works are done. Quantitively, through linear regression, Zhou et al. analyzed the relationship between various weather elements, such as temperature, humidity, wind, and other parameters, as well as dummy variables such as whether it is a holiday and whether it rains, and the deviation of passenger flow from the average value in Shenzhen[16], while Singhal et al. analyzed the case of Bnox[17], Arana et al. analyzed the case of Gipuzkoa using similar methods[18]. Zhou et al find out that negative impact of inclement weather on bus ridership appears more obvious during off-peak time, Arana et al. showed that the destination, which may imply the purpose of travel, will cause the impact of weather on passenger flow differ. Singhal et al. discovered that passenger flow in elevated stations is more likely to be affected compared with underground stations. However, most of previous works focus more on fitting than predicting, which lack of practicality and is hard to directly enhance scheduling of trains according to changes in weather.

As weather is closely related with passenger flow, it is therefore important to evaluate the effect of bringing external factors, such as weather which may play a considerable role in citizens' travel decisions into passenger flow prediction, which can provide an effective way to make the prediction more accurate. The main contributions of this paper are as follows.

(1) A statistical method of station classification that optimizes prediction accuracy in practice. Different stations are affected differently under the same abnormal weather conditions. Most of the previous results directly predicted the overall line network, or directly selected stations with larger passenger flow. We provide a method for classifying stations based on statistical characteristics of passenger flow. This classification can effectively distinguish different passenger flow time distribution rules, and to a certain extent, it also shows the different uses of the site.

(2) Contribution of different weather parameters to improving passenger flow forecast are found out. Whether adding a weather parameter improves the prediction and how much it will improve are tested and compared. Standardized weather variable such as temperature, wind speed, humidity, and rainfall, as well as their various combinations, were separately assembled with the temporal data and historical passenger flow data, station-level forecast were separately built for each scenario to comprehensively analyze the impacts of different weather conditions on hourly metro passenger flow forecasting.

Due to the limited data size and that the deep neural network predictions are black-box predictions[8], we chose bagging regressor. It is not the most advanced of all forecasting methods, but considering that our research theme is to evaluate the impact of weather on the accuracy of passenger flow forecasting, it is only necessary to obtain the relative magnitude of forecasting accuracy under the same model to show that the contribution of each weather variable, namely the type of impact on the model after the prediction model and the magnitude of the impact in different scenarios.

The remainder of this paper is organized as follows. Section 2 describes models and methods we used to predict the passenger flow. Section 3 introduced the performance of models with different weather variable inputs and provided possible explanation. Section 4 concludes the paper and suggests directions for possible improvements of the work and directions for further study.

## 2. Methods

### 2.1. Dataset

The area of study includes an Asian metropolitan from April 1 to June 29, 2018.

The dataset is divided into passenger flow data and weather data. The passenger flow data is from 6:00 to 23:00 from April 1 to June 29, 2018, which is represented as an Origin-Destination (OD) matrix. Weather data includes air temperature (°C), wind speed (km/h), humidity (%) and air pressure (mbar).

Table 1. Weather data (sample)[19]. Weather data includes air temperature (°C), wind speed (km/h), humidity (%) and air pressure (mbar).

| Date | Time | Temperature (°C) | Wind speed (km/h) | Humidity (%) | Barometer (mbar) |
|---|---|---|---|---|---|
| 2018-04-01 | 6:00 | 22 | 9 | 78 | 1015 |
| 2018-04-01 | 6:30 | 22 | 11 | 78 | 1015 |
| …… | …… | …… | …… | …… | …… |
| 2018-06-29 | 22:45 | 30 | 17 | 75 | 1004 |

Table 2. Passenger flow volume data (sample). Origin, destination and outbound time and quantity of passenger are included in the dataset.

| Date | | Outbound station number | The outbound time | The number of outbound |
|---|---|---|---|---|
| 2018/4/1 | 1 | 1 | 6:00 | 2 |
| 2018/4/1 | 2 | 3 | 6:00 | 6 |
| 2018/4/2 | 4 | 5 | 6:15 | 6 |
| …… | …… | …… | …… | …… |

Given the half an hour interval of the weather data, the average over two connected time periods is calculated. The weather data exhibited relatively low amount of noise over the three months of the study period. Moreover, we found that the models did not perform as well when using derived variables such as the difference between each data point or monthly average. As such, raw data is used for model construction, and derived variables are used to distinguish different stations. Based on statistics, the daily peak hours are set from 7:30 to 8:30 and 17:15 to 19:30.

## 2.2. Regression and analysis

Here, we present machine learning methods for short-time prediction of subway passenger flow considered in this study, including data collection, sample selection, and their limitations. Traffic features are identified from historical spatiotemporal data, and predictions are made using new data. We use classical machine learning algorithms, such as decision tree algorithm and random number algorithm. We find that the bagging regressor performs the best; as such, the results presented in this study will be based on this model unless stated otherwise.

### 2.2.1. Bagging regression

A Bagging regressor is an ensemble meta-estimator that fits each base regressor on random subsets of the original dataset and aggregates their individual predictions (either by voting or by averaging) to form a final prediction. Such a meta-estimator can be used to reduce the variance of a black-box estimator (e.g., a decision tree) by introducing randomization into its construction procedure and making an ensemble out of it.

A learning set of $L_i = (X_i, Y_i), (i = 1, \ldots, n)$ where the $Y_i$ is a numerical response.

Assume we have a procedure for using this learning set to form a predictor $\varphi(x, L)$ -- if the input is $x$ we predict $y$ by $\varphi(x, L)$.

The process of Bagging method is roughly as follows:
1. Take repeated bootstrap samples $\{L^{(B)}\}$ from L, and form $\{\psi(x, L^{(B)})\}$.

2. take $\varphi_B(\mathbf{x})$ as

$$\varphi_B(\mathbf{x}) = av_B \varphi(\mathbf{x}, L^{(B)}).$$

For detailed description, we refer to L. Bbelman[20].

### 2.2.2. Multi-layer Perceptron regressor

MLPRegressor optimizes the squared error using LBFGS or stochastic gradient descent. The model is trained iteratively; at each time step, the partial derivatives of the loss function with respect to the model parameters are computed to update the parameters. For detailed description, we refer to Su *et al*[6].

### 2.2.3. Random forest regressor

A random forest is a meta-estimator that fits several classifying decision trees on various sub-samples of the dataset and uses averaging to improve the predictive accuracy and control over-fitting. For detailed description, we refer to Tin Kam Ho[21].

### 2.2.4. Multivariate Linear Regression

To intuitively display the effect of weather on the number of passengers, we used multivariate linear regression. The linear regression aims to evaluate the contribution of weather to the change of passenger flow through the size of the goodness of fit. The linear regression takes the difference between the number of passengers and the historical average as the dependent variable, and standardized humidity, temperature, air pressure and dummy variable for rain (0 for clear moments, 1 for rainy moments, 2 for heavy rain moments), as independent variables.

## 3. Results and discussion

## 3.1. Descriptive statistics

First of all, in order to simplify the problem and make the established passenger flow model easier to be applied to more stations and rail transit systems, we divide the station according to the average value and variance of passenger flow in several working days and one day of holidays. It is divided into four types: "high mean high variance, high mean low variance, low mean low variance, low mean high variance". The mean reflects the total number of passengers in a station, while the variance reflects the distribution of traffic throughout the day. In practice, since the mean and variance have a certain linear relationship, we first divide the mean into two parts, high and low, and then select sites with high variance and low variance in the two parts.

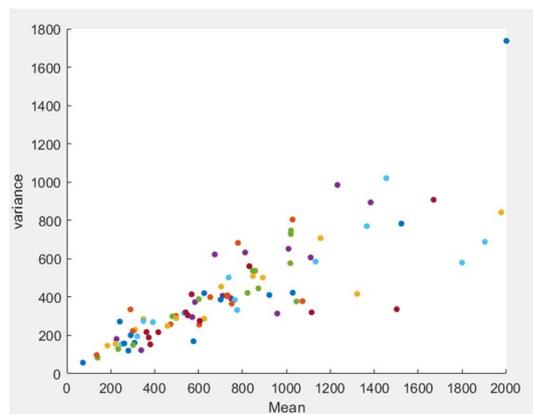

**Figure 2**. The mean-variance value distribution chart of stations. Each dot stands for a station.

The detailed site classification is as follows. In order to facilitate the subsequent analysis, we select Station 1, Station 3, Station 76, and Station 84 as the representatives of each type for analysis. The four stations represent four kinds of destination: CBD, commercial area, commuting origins or destinations and tourist attractions.[22] Their usages determined the volume and temporal distribution of passenger flow, which imply that the proposed statistical method of station classification is universal and representative in real life.

**Table 3.** Stations are classified. Considering the actual situation of data, we first divide the mean into two parts, high and low, and then select sites with high variance and low variance in the two parts. The serial numbers are those shown in Figure 1.

| The station classification | | | | | | | | | |
|---|---|---|---|---|---|---|---|---|---|
| Low average/High variance | 55 | 86 | | | | | | | |
| High average/Low variance | 3 | 6 | 15 | 75 | 76 | | | | |
| High average/High variance | 1 | 27 | 28 | 64 | 80 | 118 | | | |
| Low average/Low variance | 2 | 4 | 5 | 7 | 8 | 9 | 10 | 11 | 12 | 13 |
| | 14 | 16 | 17 | 18 | 19 | 20 | 21 | 22 | 23 | 24 |
| | 25 | 26 | 29 | 30 | 31 | 32 | 33 | 34 | 35 | 36 |
| | 37 | 38 | 39 | 40 | 41 | 42 | 43 | 48 | 49 | 50 |
| | 51 | 52 | 53 | 54 | 56 | 57 | 65 | 67 | 68 | 69 |
| | 71 | 72 | 73 | 74 | 78 | 81 | 82 | 83 | 84 | 85 |
| | 87 | 88 | 89 | 96 | 97 | 98 | 99 | 100 | 101 | 102 |
| | 103 | 114 | 115 | 116 | 117 | 119 | 120 | | | |

To evaluate the impact of weather on passenger flow and determine whether weather can effectively improve the forecast of passenger flow, we first analyze the impact of weather on passenger flow at each typical station from both qualitative and quantitative perspectives, and then use machine learning methods.

At different times of the day, or on two days of different environments, passenger flow is significantly more affected by commuting, and other mandatory factors than environmental factors such as weather. Therefore, we choose to compare the passenger flow of the same time slice in different days and divide the days into two categories – holidays and working days – to eliminate the periodic effect as much as possible.

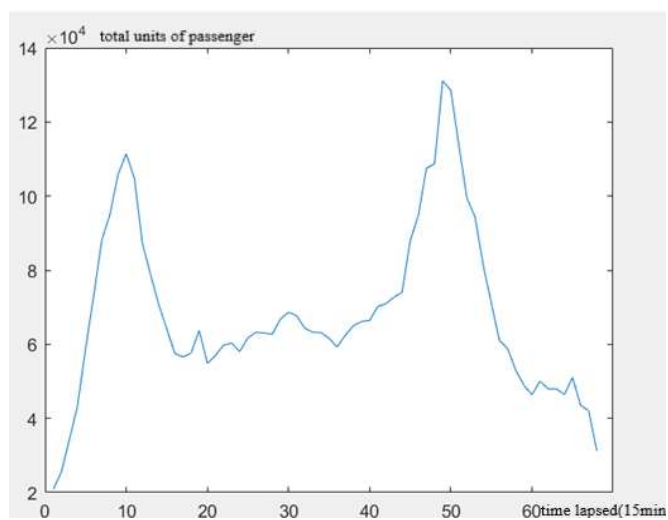

**Figure 3.** The average passenger flow of each time slice on all working days. The horizontal axis represents how many 15 minutes have passed since 6:00, and the vertical axis represents the total passenger flow of the line network.

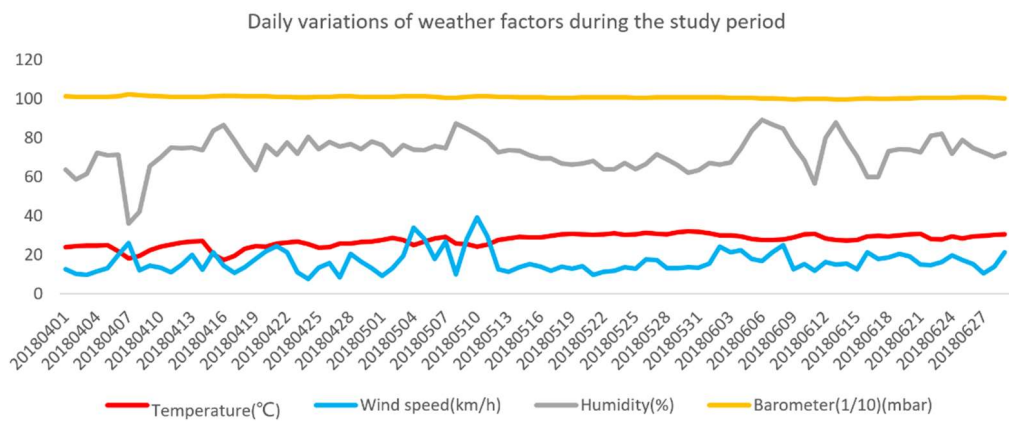

**Figure 4.** Changes in each weather indicator within the study time range.

In addition to the above indicators, rainfall is also an indicator that affects passenger travel. Considering the discreteness of the observatory data, we set the rainfall condition as a dummy variable, assigning 0 for no rain, 1 for moderate rain, and 2 for extreme weather such as heavy rain and thunderstorms.

We centralize the site traffic, that is, subtract the average traffic of the site on the date to which it belongs, to better reflect the degree to which the site deviates from the normal traffic on a certain day.

## 3.2. Model performance

After normalizing the known data (including humidity, temperature, wind, and air pressure), the data were divided into the training set and the test set at a ratio of 3:7. Then, after which the model was used to test through the test set, calculating the mean square error, root mean square error, mean absolute error, and model scores. Both the model fitting results and the three models performed similarly. For the fit of all subway stations, the Bagging Regressor and random forest regression were able to reach a score of 83, where the random forest had smaller errors. Because MLPRegressor does not perform well in the actual operation process, it is not analyzed and evaluated for error. Stations have been previously divided according to the mean value and the variance, from which 8 stations (number 55,86,1,28,3,6,4,102) are selected as representative stations. The data were divided into weekdays and weekends by date, and this particular data was fitted using a random forest. The remaining two stations still use the full data. After fitting the new station classification to the other two models, the model improved significantly in score and error. Among them, Bagging Regressor model score reached 96 and mean squared error decreased to 64101. Moreover, MLPRegressor also performs well on small-volume data, but the error is still greater than the other two models. This shows that the best Bagging Regressor performance and the least errors in the three models.

Table 4. Performance and error of different models.

| Scope of data | Model category | Mean squared error | Root mean squared error | Mean absolute error | Mode score |
|---|---|---|---|---|---|
| All the stations | BaggingRegressor | 167622.23 | 409.42 | 222.70 | 0.83 |
| | RandomForestRegressor | 160709.13 | 400.89 | 220.43 | 0.83 |
| Selected 8 stations | RandomForestRegressor (Weekend) | 82074.42 | 286.49 | 122.25 | 0.93 |
| | MLPRegressor | 72543.50 | 269.34 | 109.08 | 0.95 |
| | RandomForestRegressor (Workday) | 64327.63 | 253.63 | 102.77 | 0.95 |
| | BaggingRegressor | 64101.25 | 253.18 | 104.87 | 0.96 |

After fitting the data using the model, Bagging Regressor was finally selected as the model used in the validation work n order to analyze the influence of different weather variables on the model performance, we combined the four variables of air temperature/wind speed/humidity and air pressure and made no changes to other conventional variables. The model was established with 70% of the data and validated with 30% of the data. And the mean absolute error (MAE), mean square error (MSE) and root mean square error (RMSE) were calculated. The optimized traditional input variables include humidity, temperature, wind, and air pressure and flow data of this time slice of past 7 days.

Table 5. Results of Bagging Regressor with different variables: (a) workdays; (b) weekends.

| (a) | Input variables | | | | | Bagging Regressor | | |
|---|---|---|---|---|---|---|---|---|
| Models | Optimized Traditional Input Variables | Temperature | Wind Speed | Humidity | Barometer | MAE | MSE | RMSE |
| 1 | √ | × | × | × | √ | 77.08 | 30648.498 | 175.0 |
| 2 | √ | × | √ | × | √ | 77.13 | 30681.564 | 175.1 |
| 3 | √ | √ | √ | × | √ | 77.09 | 30727.048 | 175.2 |
| 4 | √ | √ | × | √ | √ | 77.06 | 30749.436 | 175.3 |
| 5 | √ | √ | × | × | √ | 77.04 | 30765.137 | 175.3 |
| 6 | √ | × | × | √ | × | 77.37 | 30776.441 | 175.4 |
| 7 | √ | × | √ | √ | √ | 77.16 | 30817.631 | 175.5 |

| Models | | | | | | MAE | MSE | RMSE |
|---|---|---|---|---|---|---|---|---|
| 8 | √ | × | × | √ | √ | 77.12 | 30823.987 | 175.5 |
| 9 | √ | √ | √ | × | × | 77.31 | 30845.101 | 175.6 |
| 10 | √ | × | √ | × | × | 77.53 | 30854.171 | 175.6 |
| 11 | √ | √ | × | × | × | 77.35 | 30875.088 | 175.7 |
| 12 | √ | √ | √ | √ | √ | 77.19 | 30903.037 | 175.7 |
| 13 | √ | √ | × | √ | × | 77.44 | 31071.618 | 176.2 |
| 14 | √ | √ | √ | √ | × | 77.32 | 31090.13 | 176.3 |
| 15 | √ | × | √ | √ | × | 77.68 | 31118.74 | 176.4 |

| **(b)** | | Input variables | | | | Bagging Regressor | | |
|---|---|---|---|---|---|---|---|---|
| Models | Optimized Traditional Input Variables | Temperature | Wind Speed | Humidity | Barometer | MAE | MSE | RMSE |
| 1 | √ | × | × | × | √ | 111.50 | 51208.3 | 226.29264 |
| 2 | √ | × | √ | × | × | 112.14 | 51233.5 | 226.34827 |
| 3 | √ | × | √ | × | √ | 112.17 | 51836.9 | 227.67737 |
| 4 | √ | √ | × | × | × | 112.48 | 51996.7 | 228.02793 |
| 5 | √ | × | × | √ | × | 112.73 | 52277.2 | 228.64208 |
| 6 | √ | √ | × | √ | × | 112.82 | 52346.1 | 228.79278 |
| 7 | √ | √ | × | × | √ | 112.70 | 52473.3 | 229.07061 |
| 8 | √ | √ | √ | × | × | 112.99 | 52552.5 | 229.24351 |
| 9 | √ | √ | × | √ | √ | 112.97 | 52846.4 | 229.88363 |
| 10 | √ | √ | √ | √ | × | 113.58 | 52890.4 | 229.97914 |
| 11 | √ | × | × | √ | √ | 112.84 | 52928.1 | 230.06116 |
| 12 | √ | × | √ | √ | × | 113.42 | 52993.0 | 230.2023 |
| 13 | √ | √ | √ | × | √ | 113.23 | 53135.0 | 230.51039 |
| 14 | √ | × | √ | √ | √ | 113.52 | 53446.8 | 231.18564 |
| 15 | √ | √ | √ | √ | √ | 113.94 | 53767.7 | 231.87879 |

√: The input variable has been used in the corresponding model.
×: The input variable has not been used in the corresponding model.

In the model using working day data: MAE values range from 77.04-77.68; RMSE values range from 175.06-176.40. In the model using weekend data: MAE values ranged from 111.50-113.94; RMSE values ranged from 226.29-231.87. The reason for the large difference between the two error values is that the data volume of weekday data and weekend data is different. It is difficult to train a good multivariate regression model with less data. Therefore, it is difficult to avoid the large error of the model using weekend data. of. Among the models using weekday data: The top 5 models with the smallest error all used air pressure as one of the variables, and 3 of them used air temperature, while almost none used humidity and wind speed as variables. This is also in line with the results of the correlation analysis, that is, precipitation represented by air pressure has the greatest impact on subway passenger flow, while air temperature has a relatively small impact, while wind speed and humidity have little effect.

### 3.3. Model analysis

**Table 6.** Correlation coefficient between passenger flow and different variables. Type1 stands for high average/variance stations; type2 stands for low average/high variance stations; type3 stands for low average/high variance stations; and type 4 stands for low average/variance stations (1 for belonging to the category; 0 for not belonging).

|  | Total | Workdays | Weekends |
|---|---|---|---|
| StationType1Dummy | 0.28 | 0.29 | 0.24 |
| StationType2Dummy | -0.43 | -0.43 | -0.48 |
| StationType3Dummy | 0.40 | 0.39 | 0.51 |
| StationType4Dummy | -0.26 | -0.26 | -0.28 |
| temperature | 0.091 | 0.064 | 0.180 |
| wind | -0.0016 | -0.0098 | 0.0310 |
| humidity | 0.053 | 0.052 | 0.058 |
| barometer | -0.14 | -0.13 | -0.15 |
| Flow of 1 day ago | 0.89 | 0.93 | 0.81 |
| Flow of 2 days ago | 0.83 | 0.90 | 0.75 |
| Flow of 3 days ago | 0.82 | 0.88 | 0.74 |
| Flow of 4 days ago | 0.82 | 0.88 | 0.73 |
| Flow of 5 days ago | 0.82 | 0.88 | 0.74 |
| Flow of 6 days ago | 0.88 | 0.93 | 0.74 |
| Flow of 7 days ago | 0.95 | 0.96 | 0.96 |

As a potential explanation to the results shown in Table 11, we use correlation analysis to analyze station classification, weather information, and historical data, and obtain the correlation coefficient between each variable and subway passenger flow. The results show that the correlation coefficient of air pressure is dominant in all time periods. This is because the air pressure usually becomes lower during precipitation, so the value of air pressure reflects the rainfall to a certain extent.

As a result, the correlation coefficient of air pressure is relatively high. And for the other three weather factors: in working days, the correlation coefficient of air temperature is the largest, followed by humidity, and the effect of wind speed is the smallest. During the weekend, the correlation coefficient of air temperature and wind speed increased greatly, and the correlation coefficient of air temperature rose to 0.18. At the same time, the coefficient of humidity also increased slightly. This also proves that on weekends, the weather has a greater impact on people's travel intentions, and the temperature is the most obvious. At the same time, air pressure, as another manifestation of precipitation, also has a great impact on subway passenger flow.

From the data, the stronger the correlation, the more it will contribute to the forecast. To enhance the accuracy of station-level prediction, the following multivariate linear regression are performed. For the weekday time slice, we have the following regression model.

**Table 7.** Multivariate linear regression fitting results of weekday time slice. Difference between the number of passengers and the historical average as the dependent variable, and standardized humidity, temperature, air pressure and dummy variable for rain (0 for clear moment; 1 for rainy moments; 2 for heavy rain moments) as independent variables.

| weekday evening peak (19:30) | R | R-square | Adj. R-square | RMSE | Durbin-Watson |
|---|---|---|---|---|---|
| Station1 | .327a | .107 | .025 | 142.57386 | 1.746 |
| Station3 | .288 | .083 | -.002 | 236.16568 | 1.834 |
| Station76 | .345 | .119 | .037 | 148.67756 | 1.718 |
| Station84 | .383 | .146 | .067 | 161.30562 | 1.265 |
| weekday morning off-peak (10:00) | R | R-square | Adj. R-square | RMSE | Durbin-Watson |
| Station1 | . | .113 | .031 | 42.71508 | 2.063 |
| Station3 | .254 | .065 | -.022 | 129.53264 | 1.929 |
| Station76 | .370 | .137 | .057 | 308.89973 | 2.213 |
| Station84 | .364 | .133 | .053 | 15.03952 | 2.191 |

As can be seen from the above two tables, the Durbin-Watson values in the two tables are all within the interval of 2±1.44, which can reject the hypothesis of autocorrelation of independent variables, that is, independent variables are independent of each other. However, in both cases, the adjusted R-square value is lower, and the impact value of the evening peak is weaker than the impact value of the morning peak. It is speculated that the commuting on weekdays is dominated by compulsory transportation needs such as work and education. Such needs will not be affected by weather conditions. The proportion of such forced commutes in the normal peak hours of the weekdays is slightly lower than that in the peak seasons. Below is the following regression model for holidays.

**Table 8.** Multivariate linear regression fitting results of weekday time slice. Difference between the number of passengers and the historical average as the dependent variable, and standardized humidity, temperature, air pressure and dummy variable for rain (0 for clear moments; 1 for rainy moments; 2 for heavy rain moments) as independent variables.

| Holiday 10:00 | R | R-square | Adj. R-square | RMSE | Durbin-Watson |
|---|---|---|---|---|---|
| Station1 | .241 | .058 | -.130 | 208.573 | 2.692 |
| Station3 | .260 | .068 | -.119 | 170.30593 | 2.628 |
| Station76 | .272 | .074 | -.111 | 479.37260 | 2.210 |
| Station84 | .425 | .180 | .016 | 17.82244 | 2.378 |
| Holiday 19:30 | R | R-square | Adj. R-square | RMSE | Durbin-Watson |
| Station1 | .220 | .048 | -.150 | 752.01242 | 2.611 |
| Station3 | .244 | .060 | -.136 | 497.32465 | 2.638 |
| Station76 | .382 | .146 | -.032 | 668.65303 | 2.802 |
| Station84 | .546 | .298 | .152 | 226.42650 | 1.534 |

As can be seen from the above two tables, the Durbin-Watson values in the two tables are all within the interval of 2±1.07, which can reject the hypothesis of autocorrelation of independent variables, that is, independent variables are independent of each other.

Although weather still only accounts for a small part of the impact of passenger traffic on weekends, its proportion has increased significantly compared with weekdays. Weather may affect some people to reduce their willingness to travel and thus reduce passenger flow, and it will also affect other people to change their travel mode to rail transit in which the riding experience and time are less affected by external weather. The two trades off each other, and finally show a partial balance in terms of total passenger flow.

It is usually believed that the temperature will lead to people's willingness to travel, but the research period is April to June, the temperature increases and average has not yet reached the threshold where people are unwilling to travel anymore, so the higher temperature can stimulate people on the weekend or usual travel will, so the time period of temperature is positively associated with traffic.

## 4. Conclusions

The study processes the passenger flow data of four both statistically and realistically different types of subways stations in an Asian city. Additionally, this study uses data from April to June 2018. We then further analyze contribution of different weather factors to predicting subway passengers flow as well. It predicts future passenger flow through data from known periods. Therefore, we select machine learning models as mention before in this report to carry out prediction models and compare them to find out the contribution of each weather factor.

Barometer, which usually represents precipitation contributed most to the enhancement of prediction. Other factors of weather may appear as noise data in some cases. Bringing weather data into prediction models for stations like Station84 (station of type 4), which is an 'unnecessary' destination, will have better effect on improving passenger flow forecasting.

Among weather factors, barometer negatively correlated with passenger flow. Barometer has the greatest impact on passenger flow, followed by humidity and temperature, which affects how people feel outdoors, and wind speed has the least impact. Each weather parameter had a more significant impact on passenger traffic on weekends. The more correlated, the more useful when predicting.

This study proves that weather can be used to improve passenger flow forecasting, and evaluates the improvement effects of different weather parameters on passenger flow forecasting in different scenarios, providing new ideas for improving passenger flow forecasting in subdivided scenarios. By putting suitable the current weather data into the model, we can obtain the flow prediction in a short time, and realize the dynamic adjustment of vehicle deployment, slow down the congestion in peak hours, and improve the operation efficiency of the subway system.

Our study also finds that while models with weather elements improved in most cases, the degree of improvement remained incremental mostly. The potential explanations are as follows. (1) The commuting on weekdays is dominated by compulsory transportation needs such as work and education[23]. Such needs will not be affected by weather conditions[24]. The proportion of such forced commutes in the normal peak hours of the weekdays is slightly lower than that in the peak seasons. (2) Weather may affect some people to reduce their willingness to travel and thus reduce passenger flow, and it will also affect other people to change their travel mode to rail transit in which the riding experience and time are less affected by external weather[25]. The influence of weather on passenger flow has a trade-off effect, resulting in a less significant impact of weather on the absolute number of passenger flow. (3) The previous machine learning methods based on historical flow have already achieved a relatively high accuracy, which left smaller space for possible enhancements.

Nevertheless, this study also has much to improve. For instance, the data used are not for many consecutive years, resulting in insufficient training of the machine learning model, while the time series model may be over-fitted, and there may be law of large numbers that cannot be revealed. It can also lead to the conclusion that it may not be suitable for studying the trend of long-term passenger flow changes. It is temporarily impossible to provide effective suggestions for public travel. The evaluation of different models has not been extended to mixing a variety of models to construct passenger flow prediction solutions that can be applied to specific scenarios.

# Acknowledgements

This work was supported by Touch Education Technology Inc. We acknowledge scientific support from Prof. Ma of ETH Zurich; editorial support from J. S. Lim of Harvard University; and administrative support from C. Ding of Touch Education Technology Inc.